\documentclass{article} 
\usepackage{iclr2027_conference,times}

\usepackage{hyperref}
\usepackage{url}
\usepackage[utf8]{inputenc} 
\usepackage{booktabs}       
\usepackage{amsfonts}       
\usepackage{nicefrac}       
\usepackage{microtype}      
\usepackage[dvipsnames]{xcolor}         

\usepackage{graphicx}
\usepackage{booktabs}
\usepackage{comment}
\usepackage{bm}
\usepackage{makecell}
\usepackage{multirow}
\usepackage{xspace}
\usepackage[accsupp]{axessibility}  
\usepackage{graphicx}
\usepackage{subcaption}
\usepackage{enumitem}

\makeatletter
\DeclareRobustCommand\onedot{\futurelet\@let@token\@onedot}
\def\@onedot{\ifx\@let@token.\else.\null\fi\xspace}

\def\etc{\emph{etc}\onedot} 
\def\wrt{w.r.t\onedot}

\makeatother

\usepackage{array}
\newcolumntype{C}[1]{>{\centering\arraybackslash}m{#1}}
\newcolumntype{M}[1]{>{\centering\arraybackslash}p{#1}}

\definecolor{darkgreen}{RGB}{0, 100, 0}   
\definecolor{DarkMagenta}{rgb}{0.7, 0.0, 0.7}

\title{Tokens of Motion: Primitive Motion Learning for Human-Humanoid Skill Transfer}

\author{
Hanwen Ren$^{1,2}$*,
Meng Zheng$^{1}$,
Wenzhe Cui$^{1}$,
Benjamin Planche$^{1}$,
\\
\textbf{Zhongpai Gao}$^{1}$, 
\textbf{Anwesa Choudhuri}$^{1}$,
\textbf{Terrence Chen}$^{1}$,
\textbf{Ziyan Wu}$^{1}$\\[0.3em]
$^{1}$United Imaging Intelligence, Boston, MA, USA \\
$^{2}$Purdue University, West Lafayette, IN, USA\\
}

\iclrfinalcopy 
\begin{document}

\begingroup
\renewcommand{\thefootnote}{\fnsymbol{footnote}}
\footnotetext[1]{This work was carried out during the internship of Hanwen Ren at United Imaging Intelligence, Boston, MA}
\endgroup
\setcounter{footnote}{0}

\maketitle
\lhead{} 

\begin{abstract}
Human videos provide a scalable source of manipulation experience, yet transferring them to robots remains challenging because human demonstrations lack robot-compatible actions and differ substantially from robots in embodiment and kinematics. We propose \textit{Motion Primitives for Action Composition and Transfer} (M-PACT), a framework that addresses this challenge by factorizing primitive motion learning from embodiment-specific motion realization. This separation allows us to learn generalizable and reusable motion primitives from large-scale 3D human hand-pose trajectories, while requiring only limited embodiment-specific robot data to ground these primitives into executable motions. Specifically, we first discover local motion segments from human hand trajectories and learn a compact primitive-motion representation that captures recurring manipulation patterns independently of robot embodiment. Given task and observation context, these primitives are then selected and composed, while a lightweight embodiment-specific decoder maps them to robot motions. Crucially, transferring to a new embodiment preserves the learned primitive representation and requires adapting only the realization model, substantially reducing the dependence on costly robot demonstrations. 
Experiments across EgoDex, HoloAssist, and HumanoidEveryday demonstrate cross-dataset and cross-embodiment transferability of proposed M-PACT, with lightweight decoder adaptation achieving strong humanoid motion reconstruction using as few as $\sim$100 robot demonstrations. These results establish factorized primitive learning as a scalable bridge from large-scale human motion data to humanoid manipulation.
\end{abstract}

\section{Introduction}

Developing robots capable of diverse manipulation in unstructured environments remains a central goal in robotics and artificial intelligence. Most manipulation policies are learned from robot-centric experience, including demonstrations collected through teleoperation or kinesthetic teaching and interactions acquired through reinforcement learning or autonomous exploration. Although such data provide directly executable actions, collecting them at sufficient scale and diversity remains costly and difficult across tasks, environments, and embodiments \cite{rt12022arxiv,o2024open}. Human videos offer a compelling alternative: they are abundant and capture diverse object interactions, behavioral strategies, and long-horizon activities in natural settings \cite{Ego4D2022CVPR}. They can therefore complement limited robot demonstrations and broaden the task and behavioral coverage of robot learning systems \cite{kareer2025egomimic}. However, human demonstrations lack robot-compatible action supervision, while human and robot effectors differ substantially in morphology, kinematics, appearance, and feasible action spaces \cite{kareer2025egomimic,chen2025vidbot}. Directly imitating human trajectories can consequently produce ambiguous or kinematically infeasible robot actions. Bridging this embodiment gap requires representations that abstract away embodiment-specific execution while preserving transferable motion structure,
for effective and generalizable robot control.

\begin{figure*}[!t]
    \centering
    \includegraphics[width=1\textwidth]{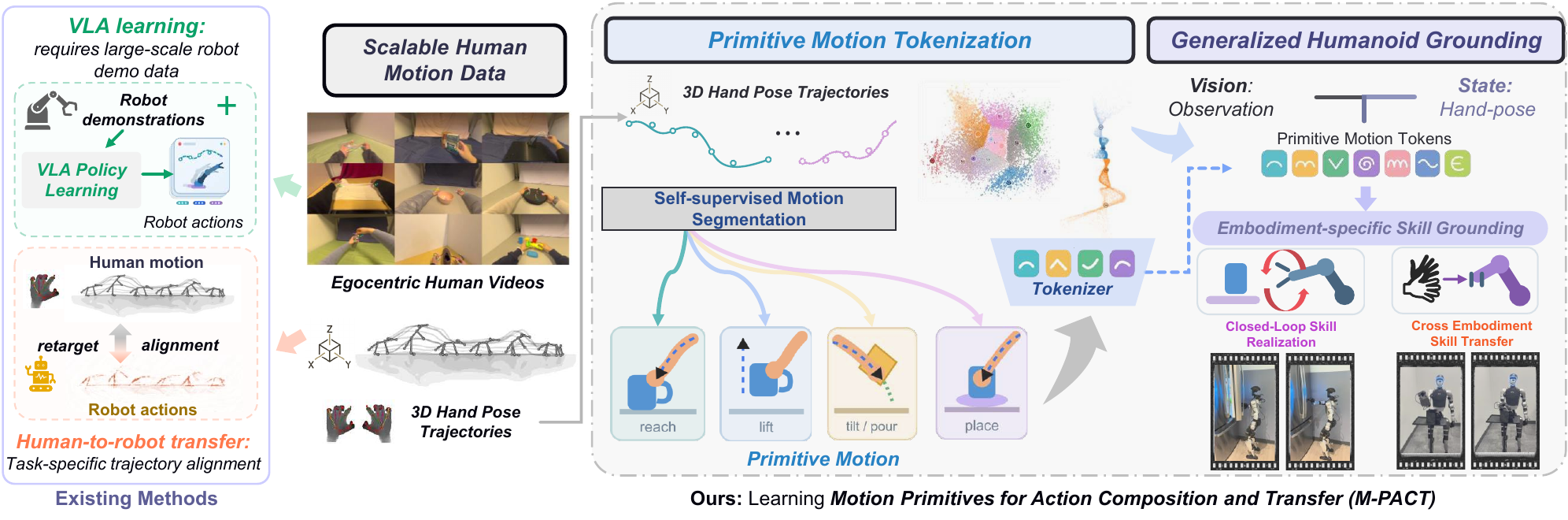}
    \caption{Proposed \textit{Motion Primitives for Action Composition and Transfer} (M-PACT) pipeline.}
    \vspace{-1.5em}
    \label{fig:teaser}
\end{figure*}

Vision-language-action (VLA) models \citep{pertsch2025fast,black2024pi_0,intelligence2025pi,liu2025rdtb,rt22023arxiv} offer a promising framework for general-purpose manipulation by mapping visual observations and language instructions directly to robot actions. Recent models such as $\Psi_0$~\cite{wei2026psi0}, H-RDT~\cite{bi2026hrdt}, and Being-H0~\cite{beingbeyond2025beingh0} further pretrain on large-scale egocentric human videos before adapting to specific robot embodiments, demonstrating the potential of human data to broaden behavioral coverage. However, two limitations remain. First, human videos lack robot-executable actions, so converting them into control supervision still requires costly robot demonstrations or computationally intensive human-motion estimation. Second, existing pretraining paradigms often couple human motion with scene appearance and context rather than explicitly isolating transferable motion structure, limiting generalization across scenes, tasks, and embodiments.

Existing human-to-robot skill transfer approaches primarily bridge the embodiment gap by geometric alignment of human hand and robot trajectories \cite{kareer2025egomimic,chen2025vidbot,shi2025zeromimic,haldar2025pointpolicy,zhou2026traj2action,okami2024}, visual transformation of human demonstrations into robot-centric observations \cite{li2025h2r,lepert2025phantomtrainingrobots,chen2026anyworld}, or reconstructing hand-object interactions and retargeting to dexterous robot hands \cite{wang2024dexcap,mu2026deximit,singh2024hoi_pretraining,chen2026dexterous,paliwal2026doasido,ObjRetarget2026}. 
However, hand-object manipulation is inherently compositional and behaviorally multimodal: a single task may comprise heterogeneous motions such as grasping, lifting, placing, \etc, each admitting multiple valid executions. Representing such a skill as a monolithic, full-trajectory supervision signal requires an imitation-learning policy to model long and highly variable motion sequences while simultaneously resolving human-robot embodiment differences, making learning difficult and limiting the reuse of shared behaviors across tasks. 
Recent approaches have begun to introduce temporal hierarchy, 
however, these methods generally transfer human behavior through complete trajectories \cite{singh2024hoi_pretraining,chen2025vidbot}, coarse semantic subtasks \cite{mu2026deximit,shi2025zeromimic}, fixed-horizon trajectory and action predictions \cite{haldar2025pointpolicy,zhou2026traj2action}. Consequently, recurring local motion patterns can remain coupled to task-level ordering and global scene context, limiting their systematic recomposition and transfer across novel tasks and environments. 

To this end, we propose to learn \textit{Motion Primitives for Action Composition and Transfer} (M-PACT), a framework that learns localized \textit{primitive motions} from human videos by factorizing transferable interaction structure from scene- and embodiment-specific realizations. This factorization allows individual primitives to be transferred across embodiments and compositionally reused across tasks and environments. At its core, a self-supervised motion segmentation module, SkillSeg, partitions 3D human hand trajectories using frame-level kinematic cues, 
and maps the resulting segments into an embodiment-agnostic vocabulary of primitive-motion tokens, by \textit{Embodied Primitive Motion Tokenization} (EPMT) module. Learned from web-scale egocentric video corpora, EPMT transforms abundant, behaviorally diverse, and cost-effective human pose trajectories into reusable manipulation primitives. 
We then condition these primitive representations on contextual features from visual foundation models and perform lightweight fine-tuning using labeled video-motion-action data, enabling M-PACT to select, compose, and realize primitives for diverse manipulation tasks. By separating reusable motion structure from task-specific context and robot-specific control, our approach substantially reduces the need for costly robot demonstrations while improving generalization across tasks and scenes. In summary, our contributions are threefold:
\begin{itemize}[leftmargin=10pt]
\item We propose to learn a motion tokenizer, EPMT that discovers localized, embodiment-agnostic primitive motions from human hand trajectories. Trained on low-cost, web-scale hand-pose data extracted from egocentric videos, EPMT learns primitive tokens that can be reused for skill transfer across embodiments.
\item We introduce M-PACT, a multi-stage transfer framework that integrates EPMT with context-aware visual representations to select and compose primitive tokens and ground them into embodiment-specific, robot-executable actions. This design supports manipulation tasks of varying horizons with lightweight fine-tuning on labeled video-motion-action data.

\item We evaluate motion reconstruction, cross-dataset representation reuse, humanoid decoder adaptation, and contextual primitive prediction across three datasets, complemented by qualitative real-world humanoid demonstrations.

\end{itemize}

\section{Related Work}

\noindent\textbf{Learning Manipulation from Human Demonstrations.}
Human demonstrations offer a scalable source of manipulation experience, motivating approaches based on shared representations, geometric alignment, and motion retargeting. EgoMimic jointly trains policies from aligned egocentric human and robot data, while Point Policy predicts keypoint trajectories that are converted into robot actions through geometric constraints \cite{kareer2025egomimic,haldar2025pointpolicy}. For dexterous manipulation, DexCap transfers captured hand motions through inverse kinematics and imitation learning, whereas DexImit reconstructs hand--object interactions and synthesizes robot demonstrations through subtask decomposition and trajectory generation \cite{wang2024dexcap,mu2026deximit}. In contrast, we first learn a compact vocabulary of local hand-motion primitives, then separately model contextual selection and embodiment-specific decoding, disentangling reusable motion structure from task context and robot execution.

\noindent\textbf{Temporal Abstraction and Reusable Skill Discovery.}
Temporal abstraction organizes demonstrations into reusable behavioral units. CompILE jointly discovers segment boundaries and latent behaviors, while Play-LMP and SPiRL learn latent plans or skill priors from unstructured experience \cite{kipf2019compile,lynch2020learning,pertsch2021accelerating}. MimicPlay learns a planner through human hand-trajectory prediction and uses it to guide a robot policy trained from teleoperated demonstrations \cite{wang2023mimicplay}. Our approach instead discovers boundaries from hand kinematics and represents each segment through factorized translation, rotation, and aperture effects. These local primitives are not predefined semantic subtasks and can therefore be reused across demonstrated skills before contextual prediction is learned over the fixed vocabulary.

\noindent\textbf{Discrete Action Representations and Motion Tokenization.}
Discrete action representations support categorical prediction and sequence modeling. VQ-BeT uses hierarchical vector quantization to model multimodal actions, while FAST tokenizes transformed robot action sequences for efficient autoregressive learning \cite{lee2024behavior,pertsch2025fast}. LAPA extends action tokenization to videos without robot labels by learning latent actions between frames and subsequently adapting with robot data \cite{ye2025latent}. Our framework instead learns from segmented 3D hand trajectories under a shared kinematic convention. Translation and aperture use independent discrete codebooks, rotation remains continuous, and the decoder reconstructs each segment's net effect conditioned on its initial state. This structure enables embodiment-specific adaptation while keeping the motion encoders and codebooks fixed.

\section{Method}
\label{sec:method}

\subsection{Framework Overview}
\label{sec:framework}
Our central design separates a shared motion representation from embodiment-specific effect decoding. We learn the representation from human hand trajectories and retain it when fitting a compact decoder to robot demonstrations. This allows us to study which motion information can be reused across datasets and embodiments, and which mappings require adaptation.

\begin{figure*}[!t]
    \centering
    \includegraphics[width=1\textwidth]{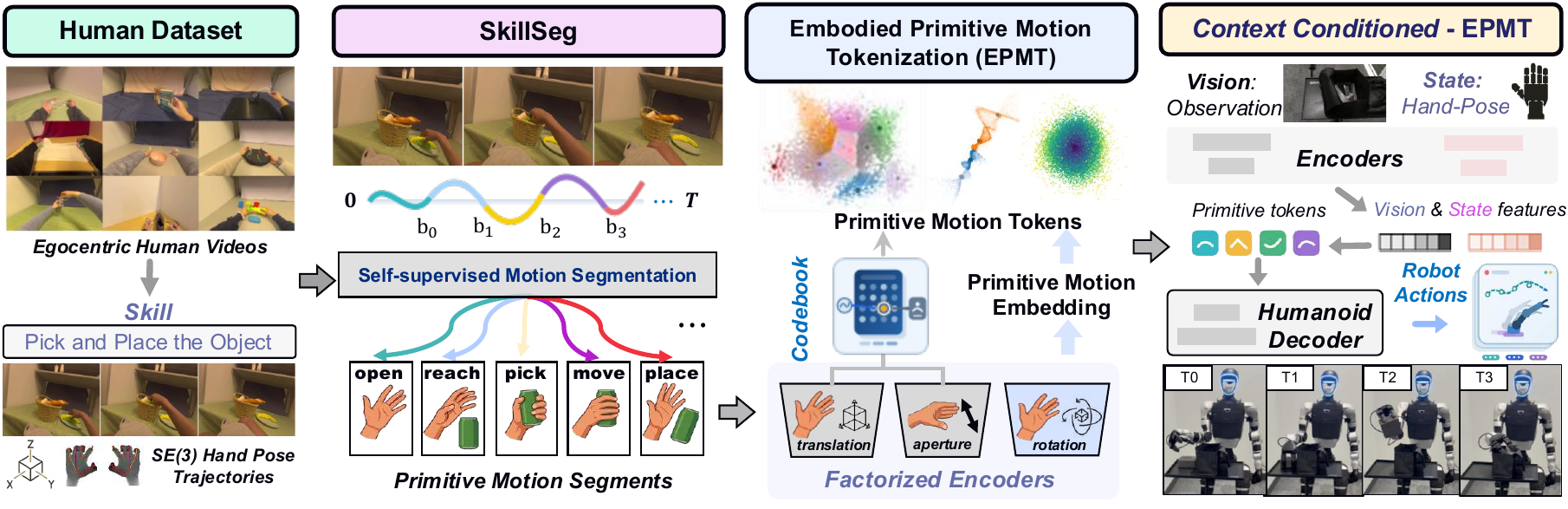}
    \caption{Overview of \textit{Motion Primitives for Action Composition and Transfer} (M-PACT) pipeline.}
    \vspace{-1em}
    \label{fig:pipeline}
\end{figure*}

The framework has three stages. SkillSeg first partitions trajectories into local motion segments using self-supervised reconstruction. \textit{Embodied Primitive Motion Tokenization} (EPMT) then encodes each segment into discrete translation and aperture tokens and a continuous rotation latent, trained jointly with a state-conditioned effect decoder. Finally, target-embodiment adaptation fits a new decoder while freezing the segmentation model, motion encoders, and codebooks. Thus, embodiment adaptation changes the mapping from a shared representation to motion effects while preserving the representation itself. An illustration of the proposed pipeline is shown in Figure~\ref{fig:pipeline}.

We study two extensions of this core. Context-Conditioned EPMT (CC-EPMT) predicts the representation from information available at a segment's start. Human-to-humanoid grounding continues the robot decoder on human-derived representations to examine the resulting reconstruction trade-off. These components are trained in stages. Representation reconstruction, contextual prediction, and offline motion realization are evaluated separately, so their respective errors can be distinguished.

\subsection{Shared Cross-Embodiment Motion Representation}
\label{sec:shared_representation}

We represent a demonstration by a hand-state sequence
$\tau=\{(p_t,R_t,\alpha_t)\}_{t=0}^{T-1}$, where $p_t\in\mathbb{R}^3$ is wrist position, $R_t\in\mathrm{SO}(3)$ is wrist orientation, and $\alpha_t\in\mathbb{R}$ is hand aperture. For human demonstrations, aperture is the thumb-index fingertip distance and poses are in the head-camera convention. Robot trajectories are recovered through forward kinematics and mapped into the same convention using a coordinate transformation, arm-length scaling, wrist-orientation alignment, and an affine aperture calibration. We denote this embodiment-specific mapping by $\mathcal A_e$ and use $(p_t,R_t,\alpha_t)$ for the aligned quantities throughout the method. See Appendix for concrete humanoid mapping.

The shared convention makes motion inputs comparable across embodiments; it does not require identical kinematics or guarantee that an aligned human motion is robot-reachable. Statistical normalization is applied separately: the tokenizer uses its source-training statistics, while each embodiment-specific decoder carries its own hand-state and effect normalizers. We write $\pi_6(R)\in\mathbb{R}^6$ for the concatenated first two columns of a rotation matrix and $\operatorname{Log}_{v}(R)\in\mathbb{R}^3$ for its axis--angle rotation vector.

\subsection{SkillSeg: Self-Supervised Motion Segmentation}
\label{sec:skillseg}
Local motion phases provide temporal units for learning reusable primitives. SkillSeg discovers their boundaries without annotated segment labels by learning a differentiable temporal decomposition that supports motion reconstruction for cross-embodiment generalization. A temporal network receives smoothed kinematic features $f_t$ containing translational velocity, speed, aperture, aperture rate, and rotation-representation derivatives. It predicts an increment $q_t\in(0,1)$. A binary motion gate $m_t$ suppresses accumulation during inactive frames, yielding a soft segment index $c_t=\sum_{s=0}^{t}m_s q_s$.
For a segment slot $k$, soft membership is defined by a triangular kernel:
\begin{equation}
    w_{t,k}=\max\left(0,1-\frac{|c_t-(k+1/2)|}{\rho}\right),
    \qquad
    \widetilde w_{t,k}=\frac{w_{t,k}}{\sum_s w_{s,k}+\varepsilon},
\end{equation}
where $\rho=0.6$. A frame encoder is pooled into a segment embedding
$a_k=\sum_t\widetilde w_{t,k}F(f_t)$, which is quantized through an auxiliary vector-quantization codebook. With $v_t$ denoting translational velocity, an auxiliary decoder reconstructs the motion target $d_k=\sum_t w_{t,k}v_t$. This membership-weighted velocity sum is proportional to a soft displacement integral at the fixed frame rate. Both the segment embeddings and targets depend on the boundary predictions, allowing reconstruction gradients to shape the temporal decomposition.

The segmentation objective combines reconstruction, vector-quantization, rate, and entropy terms:
\begin{equation}
\label{eq:skillseg_loss}
    \mathcal L_{\mathrm{seg}}
    =\mathcal L_{\mathrm{motion}}+\lambda_c\mathcal L_{\mathrm{VQ}}
    +\lambda_r\frac{(c_{T-1}-N^\star)^2}{\max((N^\star)^2,1)}
    -\lambda_e H(q).
\end{equation}
Here, $\mathcal L_{\mathrm{motion}}$ is mean squared error between the auxiliary decoder's predictions and $d_k$, averaged over active slots; $\mathcal L_{\mathrm{VQ}}$ is the auxiliary quantization loss. A slot is active when its membership mass exceeds $10^{-3}$. The duration-dependent target $N^\star$ controls the segment count, and $H(q)$ is mean Bernoulli entropy. The rate term controls temporal granularity, while the entropy term discourages saturated increments. The segmentation decoder receives no initial hand-state information. Its auxiliary encoder and codebook serve segmentation training and are distinct from the factor-specific EPMT modules below.

At inference, we average the increments of $M=7$ independently trained networks, apply the common motion gate, and extract boundaries at successive integer crossings of the cumulative score. Crossings within $K=8$ frames of the previous boundary are skipped. Including the trajectory endpoints gives $0=b_0<\cdots<b_M=T$, with segment $i$ containing frames $[b_{i-1},b_i)$. Training details and gate definitions appear in Appendix.

\subsection{EPMT: Factorized Primitive Motion Tokenization}
\label{sec:epmt}

EPMT learns recurring motion components across demonstrated skills without receiving skill labels. Separate encoders expose translation, rotation, and aperture as individually addressable channels, while a common decoder reconstructs their combined motion effect. This design provides a structured interface for factor prediction and decoder adaptation; it does not impose statistical independence between factors.

\paragraph{Segment features and factorized encoding.}
Let $a_i=b_{i-1}$ and $e_i=b_i-1$ be the first and last included frames of segment $i$. We linearly resample each channel at $L=8$ uniformly spaced points:
\begin{equation}
\begin{aligned}
(x_i^{\mathrm{tr}},x_i^{\mathrm{rot}},x_i^{\mathrm{ap}})
&=\operatorname{resample}_L\!\left(
\{(p_t-p_{a_i},\pi_6(R_{a_i}^{\top}R_t),\alpha_t)\}_{t=a_i}^{e_i}\right) \in\mathbb R^{3L}\times\mathbb R^{6L}\times\mathbb R^L.
\end{aligned}
\end{equation}
Translation is expressed as displacement from the segment start in the shared coordinate axes; rotation is relative to its initial wrist orientation. Aperture retains its absolute aligned value. Let $\bar x_i^f=\mathcal N_x^f(x_i^f)$ denote component-wise standardization using fixed source-training statistics for factor $f$. Translation and aperture are quantized, whereas rotation remains continuous:
\begin{equation}
\begin{aligned}
(z_i^{\mathrm{tr}},z_i^{\mathrm{rot}},z_i^{\mathrm{ap}})
&=\bigl(\operatorname{VQ}_{16}(E_{\mathrm{tr}}(\bar x_i^{\mathrm{tr}})),
E_{\mathrm{rot}}(\bar x_i^{\mathrm{rot}}),
\operatorname{VQ}_{4}(E_{\mathrm{ap}}(\bar x_i^{\mathrm{ap}}))\bigr)\\[-2pt]
&\in\{1,\ldots,16\}\times\mathbb R^3\times\{1,\ldots,4\}.
\end{aligned}
\end{equation}
The operator $\operatorname{VQ}_K$ returns the nearest codebook entry's index. The codebook sizes are $K_{\mathrm{tr}}=16$ and $K_{\mathrm{ap}}=4$, with 32-dimensional embeddings and lookups $e^f(k)=c_k^f$ for $f\in\{\mathrm{tr},\mathrm{ap}\}$. We distinguish the symbolic representation from its embedded decoder input:
\begin{equation}
    Z_i=(z_i^{\mathrm{tr}},z_i^{\mathrm{rot}},z_i^{\mathrm{ap}}),
    \qquad
    \widetilde Z_i=[e^{\mathrm{tr}}(z_i^{\mathrm{tr}}),z_i^{\mathrm{rot}},e^{\mathrm{ap}}(z_i^{\mathrm{ap}})]
    \in\mathbb{R}^{67}.
\end{equation}
This hybrid representation makes translation and aperture independently enumerable while retaining continuous rotational variation. The two discrete indices require six bits in total; the three-dimensional rotation latent has an additional precision-dependent storage cost. Factorization is a structural choice rather than a requirement for minimum reconstruction error, and we compare it with joint encoding in the experiments.

\paragraph{State-conditioned effect reconstruction.}
\label{sec:decoder}
The initial state and net motion effect are
\begin{align}
    h_i&=[p_{a_i},\pi_6(R_{a_i}),\alpha_{a_i}]\in\mathbb{R}^{10},\\
    y_i&=[p_{e_i}-p_{a_i},\operatorname{Log}_{v}(R_{a_i}^{\top}R_{e_i}),\alpha_{e_i}-\alpha_{a_i}]
       \in\mathbb{R}^{7}.
\end{align}
Let $\mathcal N_h$ and $\mathcal N_y$ denote component-wise standardization of states and effects, with $\bar h_i=\mathcal N_h(h_i)$ and $\bar y_i=\mathcal N_y(y_i)$. The three effect channels are denoted by superscripts $\mathrm{tr}$, $\mathrm{rot}$, and $\mathrm{ap}$. The decoder predicts a normalized effect: $\widehat{\bar y}_i=D_\theta(\mathcal N_h(h_i),\widetilde Z_i)$, $\hat y_i=\mathcal N_y^{-1}(\widehat{\bar y}_i)$.
Conditioning on $h_i$ permits a shared primitive representation to produce different effects from different initial configurations. We balance the three factors by averaging errors within each channel before averaging across channels:
\begin{equation}
    \ell_{\mathrm{bal}}(\hat{\bar y},\bar y)
    =\frac13\left(
      \frac{\|\hat{\bar y}^{\mathrm{tr}}-\bar y^{\mathrm{tr}}\|_2^2}{3}
      +\frac{\|\hat{\bar y}^{\mathrm{rot}}-\bar y^{\mathrm{rot}}\|_2^2}{3}
      +|\hat{\bar y}^{\mathrm{ap}}-\bar y^{\mathrm{ap}}|^2\right).
\end{equation}
The encoders and decoder minimize:
\begin{equation}
    \mathcal L_{\mathrm{EPMT}}
    =\mathbb E_i\left[\ell_{\mathrm{bal}}(\widehat{\bar y}_i,\mathcal N_y(y_i))
      +\beta\sum_{f\in\{\mathrm{tr},\mathrm{ap}\}}\mathcal L_{\mathrm{commit},i}^{f}\right].
\end{equation}
The commitment losses penalize encoder distance from stop-gradient codebook embeddings. Straight-through estimation passes reconstruction gradients to the discrete-factor encoders, and exponential moving averages update the codebooks. Although the encoders observe within-segment trajectories, supervision reconstructs only their net effects. Thus, the representation summarizes motion outcomes without explicitly requiring preservation of the full trajectory.

\subsection{Cross-Embodiment Decoder Adaptation and Grounding}
\label{sec:decoder_adaptation}

\paragraph{Target-embodiment adaptation.}
To transfer the learned representation to a target embodiment, we align robot demonstrations through $A_e$ and process them using the frozen SkillSeg and EPMT modules. The resulting segments provide tuples $(h_i^e,\widetilde{Z}_i^e,y_i^e)$, from which we train a new embodiment-specific decoder:
\begin{equation}
\theta_e =
\arg\min_{\theta}
\mathbb{E}_i
\left[
\ell_{\mathrm{bal}}
\left(
D_\theta
\left(
\mathcal{N}_h^e(h_i^e),
\widetilde{Z}_i^e
\right),
\mathcal{N}_y^e(y_i^e)
\right)
\right].
\label{eq:target_decoder}
\end{equation}
Each decoder uses the same 77-dimensional input, 7-dimensional output, and 27,399-parameter architecture.
Its state and effect normalizers are fitted on the target training data, while the motion-feature normalization, segmentation networks, encoders, and codebooks remain fixed.
Source and target demonstrations need not be paired: each target segment supplies its own initial state, encoded representation, and observed effect.
Thus, transferring EPMT to a new embodiment requires only learning a compact realization model rather than a new motion vocabulary.

\paragraph{Human-to-humanoid grounding.}
The adaptation above learns to realize shared primitive representations extracted from robot demonstrations.
When primitives instead originate from human demonstrations, the joint distribution of initial states and primitive representations can shift.
We therefore optionally continue the robot decoder $\theta_{\mathrm{rob}}$ on aligned human segments while keeping the EPMT representation and robot-fitted normalizers fixed:
\begin{equation}
\mathcal{L}_{\mathrm{gnd}}(\theta)
=
\mathbb{E}_{i\sim\mathcal{D}_{\mathrm{hum}}}
\left[
\operatorname{MSE}
\left(
D_\theta
\left(
\mathcal{N}_h^{\mathrm{rob}}(h_i^{\mathrm{hum}}),
\widetilde{Z}_i^{\mathrm{hum}}
\right),
\mathcal{N}_y^{\mathrm{rob}}(y_i^{\mathrm{hum}})
\right)
\right].
\label{eq:grounding}
\end{equation}
All continuation inputs and targets come from human segments; no paired human--robot trajectories or robot-effect labels are required.
Only the embodiment-specific decoder is updated, preserving the learned primitive vocabulary.
We evaluate the resulting trade-off between reconstruction of human-derived motions and preservation of native robot fidelity.
Detailed sequential replay, inverse-kinematics evaluation, and execution procedures are provided in the Appendix.

\subsection{CC-EPMT: Context-Conditioned Primitive Prediction}
\label{sec:cc_epmt}

EPMT encodes a segment after observing its motion. CC-EPMT instead predicts its representation from context available at the segment's start.

\paragraph{Context variants.}
The model accepts state-only context or additional visual and language features. With $\mathcal N_h^{\mathrm{ctx}}$ denoting the predictor's state standardization, representative inputs are
\begin{equation}
    u_i^{\mathrm{state}}=\mathcal N_h^{\mathrm{ctx}}(h_i),
    \qquad
    u_i^{\mathrm{full}}=[\bar f_{\mathrm{img}}(I_i),\bar f_{\mathrm{text}}(\ell_i),\mathcal N_h^{\mathrm{ctx}}(h_i)].
\end{equation}
Here, $I_i$ is the initial image, $\ell_i$ is the task name, and the image and text features come from frozen CLIP encoders. Subsets of these blocks define the text-only and text--state variants. Each context block is standardized using its training statistics. The decoder input separately uses the state normalizer associated with its fixed checkpoint.

\paragraph{Prediction objective.}
A shared trunk predicts categorical distributions $P_\phi^{\mathrm{tr}}(\cdot\mid u_i)$ and $P_\phi^{\mathrm{ap}}(\cdot\mid u_i)$ and a continuous rotation latent $\hat z_i^{\mathrm{rot}}=g_\phi(u_i)$. The frozen EPMT encoder supplies the training labels. In addition to classification and latent regression, we supervise the effect implied by the predicted factors. For each discrete factor, define the expected embedding $\mu_i^f=\sum_{k=1}^{K_f}P_\phi^f(k\mid u_i)c_k^f$, $f\in\{\mathrm{tr},\mathrm{ap}\}$.
The frozen decoder combines the expected embeddings and predicted rotation latent with the initial hand state, $\widehat{\bar y}_i^{\mathrm{soft}}
    =D_\theta\left(\mathcal N_h(h_i),
      [\mu_i^{\mathrm{tr}},\hat z_i^{\mathrm{rot}},\mu_i^{\mathrm{ap}}]\right)$.
For vectors in $\mathbb R^d$, define $\operatorname{MSE}(a,b)=d^{-1}\|a-b\|_2^2$. The contextual objective is
\begin{align}
    \mathcal L_{\mathrm{CC}}=\mathbb E_i\big[&
      -\log P_\phi^{\mathrm{tr}}(z_i^{\mathrm{tr}}\mid u_i)
      -\log P_\phi^{\mathrm{ap}}(z_i^{\mathrm{ap}}\mid u_i)\nonumber\\
      &+\lambda_{\mathrm{rot}}\operatorname{MSE}(\hat z_i^{\mathrm{rot}},z_i^{\mathrm{rot}})
      +\lambda_{\mathrm{pose}}\operatorname{MSE}(\widehat{\bar y}_i^{\mathrm{soft}},\mathcal N_y(y_i))\big].
\end{align}
Expected embeddings make the decoded-effect term differentiable \wrt the categorical predictions. The representation and decoder remain frozen; only the contextual predictor is optimized. The term couples factor prediction through the frozen decoder during training. At inference, discrete predictions use codebook entries rather than expected embeddings. We evaluate the contribution of decoded-effect supervision by setting $\lambda_{\mathrm{pose}}=0$ with the architecture and training budget fixed.

To isolate robot translation prediction, the experiments additionally fit independent translation classifiers with cross-entropy using state, text, or both. These classifiers use the frozen translation vocabulary but have no rotation/aperture heads or decoded-effect objective. They are distinguished from the shared-trunk CC-EPMT model and from the multi-network controller implementation in Appendix.

\subsection{Primitive Composition and Closed-Loop Skill Realization}
\label{sec:skill_realization}

The components above define a common interface from human motion to
embodiment-specific execution. A manipulation skill is represented as a
sequence of local primitive decisions rather than as a single retargeted
trajectory. At each decision step, CC-EPMT uses the current observation,
task context, and hand state to predict the next factorized primitive representation $\hat{Z}_t =
(\hat{z}^{\mathrm{tr}}_t,
 \hat{z}^{\mathrm{rot}}_t,
 \hat{z}^{\mathrm{ap}}_t)$.
The embodiment-specific decoder then realizes this shared representation
from the current robot state:
\begin{equation}
\hat{y}^{e}_t =
(\mathcal{N}^{e}_y)^{-1}
\left[
D_{\theta_e}
\left(
\mathcal{N}^{e}_h(h^{e}_t),
\widetilde{\hat{Z}}_t
\right)
\right].
\label{eq:closed_loop_realization}
\end{equation}
The decoded effect specifies the next embodiment-specific motion target,
which is passed to robot execution stack. After execution, the robot is
re-observed and the updated state conditions next primitive decision.
Repeating this predict--realize--observe cycle composes local primitives into
multi-step manipulation skills while allowing execution to adapt to the
state reached by previous actions.

This interface also defines cross-embodiment skill transfer in M-PACT:
human demonstrations provide the shared primitive vocabulary and, when
available, primitive sequences, while embodiment adaptation changes only
the mapping from these primitives to executable robot motion. Consequently,
the same primitive representation can support recorded human-to-humanoid
transfer as well as autonomous closed-loop primitive prediction.
Sec.~\ref{sec:real_world} evaluates these modes on the real humanoid, while
detailed motion realization, sequential replay, inverse-kinematics
evaluation, and execution procedures are provided in
Appendix.

\section{Experiments}
\label{sec:experiments}
We separately evaluate representation transfer, design choices, contextual prediction, and humanoid realization to isolate representation, prediction, and execution errors. Experimental setups are:

\noindent\textbf{Datasets.} 
EgoDex~\cite{egodex} provides representation training (27,411 episodes; 2.92M retained segments), HoloAssist~\cite{wang2023holoassist} frozen transfer (35,087; 0.80M), and HumanoidEveryday~\cite{zhao2025humanoid} G1 adaptation (4,064; 96,241). We remove segments outside 5--90 frames or with negligible change in all factors. EgoDex uses a 90/10 episode split (292,806 test segments); robot decoder evaluation uses 85/15 (14,754 test segments), and robot prediction uses 70/30 (28,871 test segments). Candidate sets average three episode-level calibration/test partitions. Task generalization holds out 37/124 task names under three seeds against matched episode splits. Statistics are training-only and the source tokenizer stays frozen.

\noindent\textbf{Evaluation Metrics.} We report median endpoint translation, rotation, and aperture error; robot translation is scaled to human-equivalent units by $\sigma=0.561$. Parentheses denote error divided by median motion magnitude (below one beats zero change). We also report code use, top-1 accuracy, and 90\%-coverage translation-set size. Replay uses interpolated-path and geodesic orientation errors; multi-run values average runs unless stated otherwise.

\subsection{EPMT Reconstruction, Transfer, and Ablations}
\label{sec:exp_epmt}
\medskip\noindent\textbf{Transfer and adaptation.}
Table~\ref{tab:exp_epmt} evaluates whether the EPMT vocabulary learned on EgoDex remains useful across datasets and embodiments. With the source decoder, EPMT reconstructs EgoDex motions with normalized errors well below the zero-change baseline. When transferred to HoloAssist without decoder adaptation, all codes remain active and normalized errors remain below one, although the increased translation and rotation errors and reduced effective vocabulary indicate a distribution shift. On HumanoidEveryday, refitting only the target decoder yields low normalized translation and rotation errors while preserving full code utilization, supporting reuse of the learned vocabulary for a different embodiment. Aperture remains challenging: despite a small absolute error of 0.229~cm, its normalized error of 1.388 does not outperform zero change because aperture variations in this dataset are limited.

\begin{table}[h!]
\centering\small
\caption{Reconstruction with a shared EPMT representation. HoloAssist uses the frozen source decoder; HumanoidEveryday fits only a target decoder. Parentheses give error relative to median motion magnitude. All discrete codes remain active across datasets.}
\label{tab:exp_epmt}
\setlength{\tabcolsep}{4pt}
\resizebox{.75\textwidth}{!}{%
\begin{tabular}{lccc}
\toprule
Metric & EgoDex & HoloAssist & HumanoidEveryday\\
\midrule
Decoder & Source & Frozen & Refitted\\
Translation (cm) & 2.658 (0.532) & 3.123 (0.811) & 3.442 (0.454)\\
Rotation ($^\circ$) & 0.931 (0.041) & 2.732 (0.152) & 0.901 (0.062)\\
Aperture (cm) & 0.439 (0.493) & 0.454 (0.525) & 0.229 (1.388)\\
Translation codes used & 16/16 & 16/16 & 16/16\\
Aperture codes used & 4/4 & 4/4 & 4/4\\
Effective translation vocabulary & 10.23 & 8.63 & 13.19\\
\bottomrule
\end{tabular}}
\vspace{-0.5cm}
\end{table}

\medskip\noindent\textbf{Ablations.}
To examine how temporal segmentation affects primitive-motion learning, we compare fixed-window, heuristic, and learned frontends through their downstream reconstruction performance.
With the same tokenizer, learned segmentation gives the lowest translation (3.022 cm) and aperture (4.514 mm) errors (Table~\ref{tab:exp_seg}). Because frontends define different targets, this tests the full segmentation--tokenization pipeline; the learned row is a single-seed precursor to final SkillSeg. Additional EPMT representation ablations are provided in the supplementary.


\begin{table}[h!]
\centering\small
\caption{Segmentation and downstream reconstruction on each frontend's own segments.}
\label{tab:exp_seg}
\begin{tabular}{lrrrr}
\toprule
Segmentation & Rate (s$^{-1}$) & Segments & Trans. (cm) & Ap. (mm)\\
\midrule
Fixed windows, 19 frames & 1.616 & 51,337 & 3.451 & 4.655\\
Previous heuristic & 1.742 & 55,353 & 3.188 & 5.220\\
Learned candidate & 1.752 & 55,669 & 3.022 & 4.514\\
\bottomrule
\end{tabular}
\end{table}

\subsection{CC-EPMT Prediction}
\label{sec:exp_cc}
We evaluate shared-trunk CC-EPMT on human motion and independent target-domain translation classifiers on robot motion.

\medskip\noindent\textbf{Human-motion prediction.}
To evaluate contextual primitive selection while separating prediction error from reconstruction error, we compare motions decoded from predicted factors with those reconstructed from recorded EPMT factors. Both use the same frozen decoder and are evaluated on 292,806 held-out EgoDex segments. All prediction variants share the same architecture and training budget; results are reported for one seed. Results from Table~\ref{tab:exp_decoded} show context improves translation accuracy from 36.37\% to 40.42\% and error from 5.307 to 4.952 cm, but remains far from recorded factors, especially in rotation ($20.28^\circ$ versus $0.93^\circ$; Table~\ref{tab:exp_decoded}). 
Prediction is therefore the main downstream error source. Removing decoded-effect supervision changes errors by at most $0.03^\circ$ (0.13 accuracy points); one seed cannot establish an auxiliary-loss benefit.

\begin{table*}[h!]
\centering\small
\caption{Human and robot contextual primitive prediction. Robot calibrated-set statistics average three partitions.}
\begin{subtable}[t]{0.53\textwidth}
\centering
\caption{CC-EPMT on held-out EgoDex; accuracy is percent.}
\label{tab:exp_decoded}
\resizebox{\linewidth}{!}{%
\begin{tabular}{lrrrrr}
\toprule
Representation & Trans. & Rot. & Ap. & Trans. & Ap.\\
& (cm) & ($^\circ$) & (mm) & top-1 & top-1\\
\midrule
Recorded EPMT & 2.658 & 0.93 & 4.39 & --- & ---\\
State-only prediction & 5.307 & 21.65 & 9.19 & 36.37 & 57.54\\
Context prediction & 4.952 & 20.28 & 8.46 & 40.42 & 59.07\\
Context, no effect loss & 4.960 & 20.25 & 8.47 & 40.29 & 59.08\\
Training-majority label & --- & --- & --- & 35.01 & 54.67\\
\bottomrule
\end{tabular}}
\end{subtable}\hfill
\begin{subtable}[t]{0.44\textwidth}
\centering
\caption{Robot translation prediction (T: text; S: state).}
\label{tab:exp_robot}
\setlength{\tabcolsep}{3pt}
\scalebox{0.92}{%
\begin{tabular}{lrrr}
\toprule
Predictor & Top-1 (\%) & Size & Cov. (\%)\\
\midrule
Frozen human (T+S) & 7.48 & 13.25 & 89.6\\
Robot (S) & 30.64 & 9.12 & 89.9\\
Robot (T) & 22.83 & 11.36 & 90.0\\
Robot (T+S) & 33.98 & 8.35 & 89.9\\
Majority & 22.56 & --- & ---\\
\bottomrule
\end{tabular}
}
\end{subtable}
\vspace{-0.5cm}
\end{table*}

\medskip\noindent\textbf{Robot prediction and candidate sets.}
To evaluate whether human-derived motion primitives can be selected from robot-domain context, we train translation classifiers conditioned on state, language, or both. These models predict primitive codes directly, without images or decoder execution. Top-1 accuracy is evaluated on 28,871 test segments using one seed, while candidate-set results are averaged over three episode-level calibration/test partitions.

Combining language and state achieves 33.98\% accuracy, compared with 30.64\% for state alone, 22.56\% for the majority baseline, and 7.48\% for frozen human transfer (Table~\ref{tab:exp_robot}). At approximately 90\% coverage, target-domain adaptation reduces the mean candidate-set size from 13.25 to 8.35 out of 16 codes. However, language reduces held-out-task accuracy by 0.8 percentage points, providing limited evidence of improved task generalization. Coverage measures inclusion of the recorded code rather than feasibility or task success; candidate-set construction is detailed in the supplement.


\subsection{Real World Experiments}
\label{sec:real_world}
We evaluate our framework on a real humanoid through four increasingly challenging settings: a) EPMT -- Motion Reconstruction. We compress a previously executed robot manipulation episode into discrete primitive tokens and decode them back into robot motion, testing whether real manipulation can be faithfully represented through a compact token space, b) Human2Robot -- Cross-Embodiment Transfer. Primitive decisions are extracted from egocentric human demonstrations and executed by the humanoid, with only the decoder adapted, testing whether human-derived motion primitives can transfer across embodiments, c) CC-EPMT (State-Only) -- Closed-Loop Token Prediction. Rather than providing the token sequence, the robot predicts the next primitive from its current arm state, executes it, and re-observes, enabling autonomous closed-loop primitive composition without visual or task context, d) CC-EPMT (Full) -- Context-Conditioned Manipulation. We further condition primitive prediction on the robot's camera observation and task instruction, testing whether visual and task context improves the selection and execution of appropriate motion primitives.

Together, these four settings progressively validate EPMT from motion reconstruction and cross-embodiment transfer to autonomous closed-loop prediction and context-conditioned manipulation. The results demonstrate how the same primitive token representation can bridge human motion and real-world humanoid execution across different levels of autonomy and context. Qualitative results can be found in \url{https://tinyurl.com/ykxkbu3z} and supplementary.

\section{Conclusion}
We presented M-PACT, a framework that bridges human demonstrations and humanoid manipulation through a shared primitive motion representation. By separating reusable motion structure from embodiment-specific realization, EPMT enables primitive reuse across datasets and adaptation to humanoid motion through a lightweight decoder, while CC-EPMT extends these primitives to context-conditioned prediction and closed-loop execution. Experiments across human and humanoid datasets, together with real-world humanoid demonstrations, validate the framework from motion reconstruction and cross-embodiment transfer to autonomous manipulation. Overall, our results demonstrate the potential of factorized primitive motion tokens as a reusable interface between scalable human motion data and embodiment-specific robot control.
\bibliography{iclr2027_conference}
\bibliographystyle{iclr2027_conference}

\end{document}